\documentclass[runningheads]{llncs}
\usepackage{graphicx}
\usepackage{amsfonts}
\usepackage{booktabs}
\usepackage{amsmath}
\usepackage{subfig}
\usepackage{multirow}
\usepackage[symbol]{footmisc}

\begin{document}

\title{Facial Expression Restoration Based on Improved Graph Convolutional Networks}
%
%
%
%
%
\author{Zhilei Liu\inst{1} \and Le Li\inst{1} \and Yunpeng Wu\inst{1} \and Cuicui Zhang\inst{2}\thanks{Corresponding author.}}
\institute{College of Intelligence and Computing, Tianjin University \\ \email{\{zhileiliu,le\_li,wuyunpeng\}@tju.edu.cn} \and School of Marine Science and Technology, Tianjin University \\
\email{cuicui.zhang@tju.edu.cn}}
\authorrunning{Zhilei Liu et al.}
\titlerunning{Facial Expression Restoration Based on IGCN}

\maketitle              

\begin{abstract}
Facial expression analysis in the wild is challenging when the facial image is with low resolution or partial occlusion. Considering the correlations among different facial local regions under different facial expressions, this paper proposes a novel facial expression restoration method based on generative adversarial network by integrating an improved graph convolutional network (IGCN) and region relation modeling block (RRMB). Unlike conventional graph convolutional networks taking vectors as input features, IGCN can use tensors of face patches as inputs. It is better to retain the structure information of face patches. The proposed RRMB is designed to address facial generative tasks including inpainting and super-resolution with facial action units detection, which aims to restore facial expression as the ground-truth. Extensive experiments conducted on BP4D and DISFA benchmarks demonstrate the effectiveness of our proposed method through quantitative and qualitative evaluations.
\end{abstract}
\vspace{-3mm}
\section{Introduction}
\begin{figure}[t]
    \centering
    \includegraphics[scale=0.57]{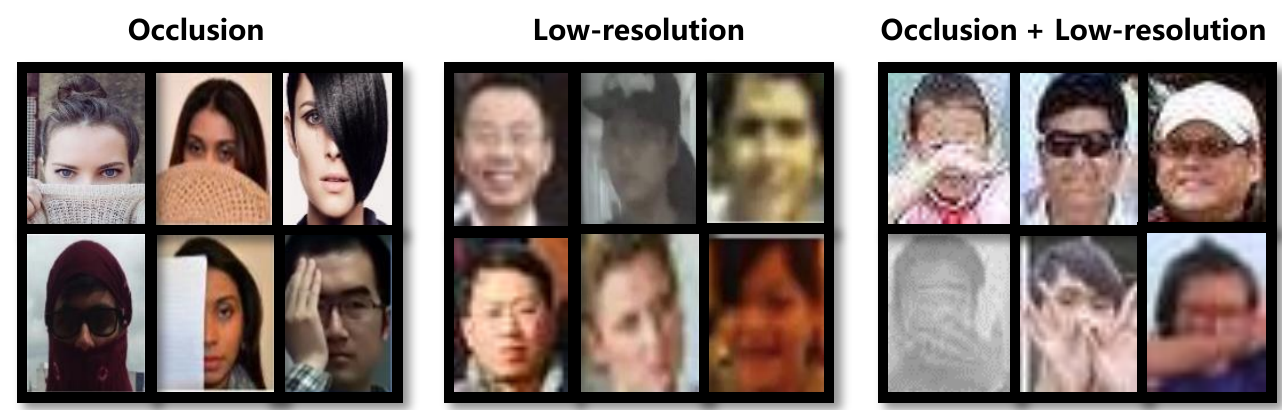}
    \vspace{-3mm}
    \caption{Example images showing face images with low resolution, partial occlusion and both in the wild.} 
    \label{fig:show}
\vspace{-7mm}
\end{figure}

\vspace{-3mm}
Facial restoration aims to recover the valuable missing information of faces caused by low resolution, occlusion, large pose, etc, which has gained increasing attention in the field of face recognition, especially with the emergence of convolution neural networks (CNN)~\cite{simonyan2014very,krizhevsky2012imagenet} and generative adversarial networks (GAN)~\cite{goodfellow2014generative}. Many sub tasks of facial restoration have achieved great breakthroughs, consisting of face completion~\cite{li2017generative,yeh2017semantic}, face super-resolution or hallucination~\cite{chen2018fsrnet,ledig2017photo}, and face frontal view synthesis\cite{huang2017beyond}. Most of these previous works just conduct these face restoration tasks independently, and what's more, only facial identity restoration is considered, without taking facial expression information restoration into consideration. Recently, some studies~\cite{song2018joint} try to jointly deal with these ill situations with the help of deep learning and GAN. And three kinds of ill facial images is shown in Fig.~\ref{fig:show} including both low resolution and occlusion faces, which give us the necessity to address both low resolution and occlusion jointly. While face super-resolution, the model takes more relations on intra-patches into account, but the relations on inter-patch during face inpainting. Concerning that graph can have intra-patch and inter-patch relationships with edges, we attempt to jointly address low resolution and partial occlusion.

Facial expression restoration is beneficial to the study of facial emotion analysis, which is easily affected by challenging environment, i.e. low resolution, occlusion, etc. In the field of facial expression analysis, facial action units (AUs) refer to a unique set of basic facial muscle actions at certain location defined by Facial Action Coding System (FACS)~\cite{ekman1997face}, which is one of the most comprehensive and objective systems for describing facial expressions. Considering facial structure and patterns of facial expression are relatively fixed, it should be beneficial for facial expression restoration if taking their relations of different AUs into consideration under occlusion and low resolution situations. However, in literature it is rare to see such facial expression restoration study by exploring the relations of different facial regions under different facial expressions.

In this paper, we propose a novel facial expression restoration framework by exploiting the correlations of different facial AUs. Our idea appears that firstly restoring the whole face and then detecting AUs to verify whether the facial expression appears or not, which can be called facial expression restoration. 
In order to learn the features of facial occluded patches from other unoccluded patches, we propose an improved graph convolutional network (IGCN), of which the structure is shown in Fig.~\ref{fig:igcn_rrmb}. In addition, IGCN also can help to improve the resolution of unoccluded face patches from other visible face patches, by exploring the correlations of different facial components. In the next, with the help of the proposed IGCN, a Region Relation Modeling Block (RRMB) is built to capture the facial features with different scales for face restoration. Given more finer facial division, more accurate relations of different face patches can be built with the help of our proposed framework. With more accurate adjacency matrix in the proposed IGCN, our proposed model can well restore the feature map in deep networks with visible patches' features. We can also use IGCN to build AU detector by exploring these correlations among AUs to help generator to restore more accurate facial expressions. Last but not least, a discriminator is designed to help generator to generate realistic faces, and additional perceptual loss is helpful to improve the quality of the generated faces to some extent. 

The contributions of this paper are threefold. First, a novel end-to-end facial expression restoration framework is proposed by jointly addressing face in-painting and face super-resolution. 
Second, an IGCN is proposed for facial patch relation modeling, and a RRMB is built with the aid of the proposed IGCN. Third, we exploit facial action units detector in generative model to improve the facial expression restoration capability of the generator.

\vspace{-3mm}
\section{Related Works}
\vspace{-3mm}
Our proposed framework is closely related to existing image restoration methods, facial action units detection methods, and graph convolutional network related studies, since we aim
to study facial expression restoration by modeling AU relations with the aid of GCN based methods.

\subsection{Image restoration}
Recently, image restoration has attracted more attentions due to the existence of generative adversarial network(GAN)~\cite{goodfellow2014generative}, which generates an image from a random vector and uses a discriminator to distinguish the real image from generated image. In order to improve the generated images' quality, many works use perceptual loss to supervise model learning~\cite{cheon2018generative}. Also, conditional GAN is proposed to limit the generated images' distribution~\cite{mirza2014conditional}. Image restoration consists of image super-resolution or hallucination~\cite{chen2018fsrnet,ledig2017photo}, image completion~\cite{li2017generative,yeh2017semantic}, face frontal view synthesis~\cite{huang2017beyond}, image denoise~\cite{muhammad2018image} and etc, 
image derain~\cite{wang2018rain}, image dehaze~\cite{engin2018cycle}, image deblurring~\cite{li2018learning} and shadow removal\cite{wang2018stacked}.
Many topics lack the true datasets. For image super-resolution, it usually uses bicubic interpolation to synthesis low-resolution image as \cite{ledig2017photo}. For image completion, it usually uses a binary mask to synthesis masked image as \cite{yeh2017semantic} and etc. We follow this setting for ill image synthesis in this paper. 
Also,~\cite{song2018joint} tries to jointly deal with face hallucination and deblurring. The situation of both problems co-occurring is common in the wild environment, which is depicted in Fig.~\ref{fig:show}. Here, we try to jointly deal with face inpainting and super-resolution for facial expression restoration concerning on relations on inter and intra patches.
Considering that face image has structure information, \cite{chen2018fsrnet} achieves amazing success with the aids of face parsing and facial landmarks information, which motivate us to restore face with facial action units information. 

\subsection{Action units detection}
Automatic facial AUs detection plays an important role in describing facial actions. To recognize facial action units under complex environment, many works have been devoted to explore various features and classifier. \cite{wu2016constrained} jointly detects facial action units and landmarks, which want to recognize AUs with the help of landmarks. \cite{khorrami2015deep} uses convolutional networks to capture deep representations to recognize AUs via deep learning model. \cite{taheri2014structure} designs AUs-related region, Zhao etal.\cite{zhao2015joint} exploits patches centered at facial landmarks, Li etal.\cite{li2017eac} exploits hand-crafted heatmaps centered at facial landmarks based on Manhattan distance and \cite{zhao2016deep} proposes deep region layer to help detect AUs. These defined regions help attention these regions while model learning. \cite{peng2018weakly} also uses AUs co-existence relations to help recognize AUs.
These methods achieve good performances for AUs detection, which motivate us to exploit the relations among AUs-related face patches. With the relations, improved graph convolutional network well fuses the features of different face patches to detect AUs, whcih is the supervisory information to help restore facial expression.  

\subsection{Graph convolutional networks}
Recently, there has been a rich line of research on graph neural networks~\cite{gilmer2017neural}.~\cite{kipf2016semi} proposes graph convolutional networks (GCN), which is inspired by the first order graph Laplacian methods. GCN mainly achieves promising performance on graph node classification task. Adjacency matrix is defined by the links between nodes of the graph. The transformation on nodes is linear transformation without learning trainable filters.
Using the relations among nodes, graph convolutional networks can embed the features of relational nodes and itself. 
We improve conventional graph convolutional network with the tensor-inputs and standard convolutional layer instead of linear transformation in conventional graph convolutional network.
\vspace{-3mm}
\section{Facial Expression Restoration based on IGCN}
\vspace{-3mm}
In this section, the proposed model is introduced in Sec.~\ref{proposed model} at first. Then, the details of improved graph convolutional networks (IGCN) are explained in Sec.~\ref{IGCN} and the region relation modeling block (RRMB) is explained in Sec.~\ref{fig:igcn_rrmb}.

\subsection{Proposed Model} \label{proposed model}
\begin{figure*}[t]
    \centering
    \includegraphics[scale=0.45]{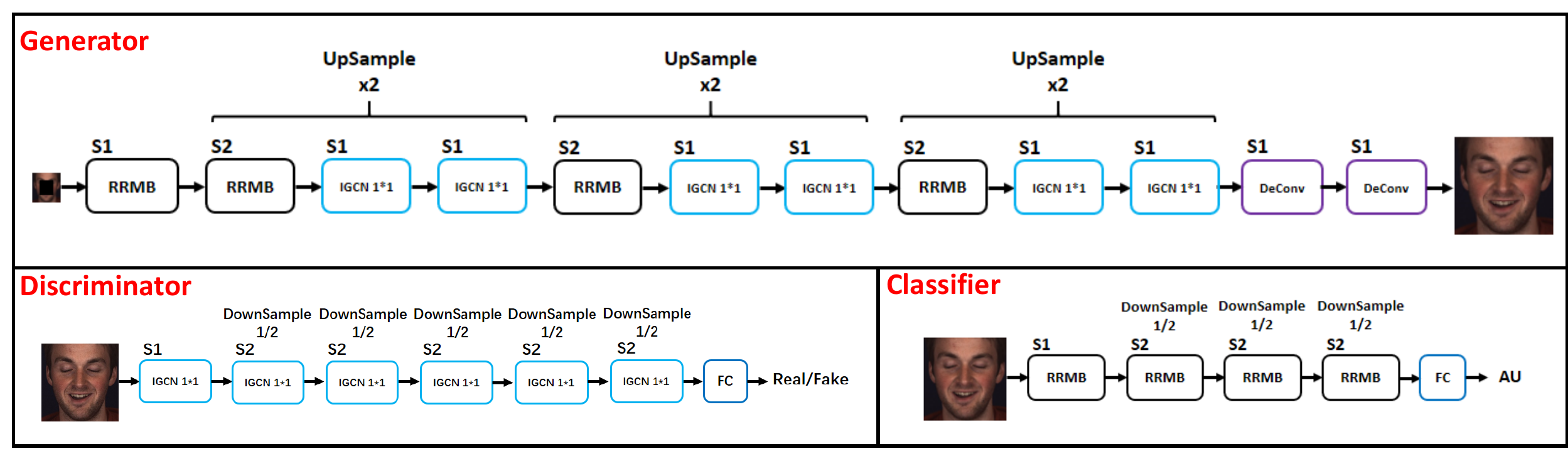}
    \caption{Framework of the proposed facial expression restoration network. During training, generator generates restored facial expression images with the supervision of pre-trained AU classifier and the adversarial loss from discriminator. During testing, only the generator is adopted to restore the ill facial images.} 
    \label{fig:proposed model}
\vspace{-3mm}
\end{figure*}

The structure of our proposed method is shown in Fig.~\ref{fig:proposed model}, which consists of a generator to restore the whole face image, a discriminator to justify whether the generated face is real or fake, a classifier to recognize the facial action units (AUs) to supervise the generated face image. To make full use of unoccluded face patches, we jointly deal with face completion and super-resolution problem. For face completion, it models the relations between unoccluded face patches and occluded patches. For face super-resolution, it models the relations among different unoccluded face patches. This aims to ensure the global harmony of generated face images.

The losses of our proposed model consist of three parts: the loss for generator learning, the loss for discriminator learning, the loss for AUs classifier learning. For generator, to help learn the generator network, we use pixel loss $L_{pix}$, which is defined as
\begin{equation}
L_{pix} = \sum \mathbb{E}_{I_{gt}, I_{out}} \left [ \left \| I_{gt}-I_{out} \right \|_2 \right ],
\label{equation: pix loss}
\end{equation}
where $I_{out}$ is the generated facial expression image by generator, $I_{gt}$ is the ground-truth facial image. Besides, we use the pre-trained 19-layer VGG to compute perceptual loss $L_{per}$ to gain more facial details~\cite{simonyan2014very}. The perceptual loss $L_{per}$ is defined as
\begin{equation}
L_{per}= \sum \mathbb{E}_{I_{gt} , I_{out}} \left[ \left\|f^{2,2}_{I_{gt}}-f^{2,2}_{I_{out}}\right\|_2 + \left\|f^{5,4}_{I_{gt}}-f^{5,4}_{I_{out}}\right\|_2 \right],
\end{equation}
where $f_{I_{gt}}^{i,j}$ is the ground-truth's feature map obtained by the $j$-th convolution layer before the $i$-th max-pooling layer in VGG-19, and $f_{I_{out}}^{i,j}$ is the generated face's.
Adversarial loss is used to improve the quality and reality of generated face image after restoration, the loss of discriminator $L_{adv}$ is defined as
\begin{equation}
L_{adv}= \sum \mathbb{E}_{I_{out} \sim p \left(I_{out} \right)}  \left[ \log D_{I} \left ( I_{out}\right ) \right ] + \mathbb{E}_{{I}_{gt} \sim p({I}_{gt})} \left[ \log \left ( 1 - D_{I} \left ( {I}_{gt}\right ) \right ) \right ],
\end{equation}
where $D_I$ is the discriminator to discriminate the ground-truth face from the generated face. In order to retain the facial expression, AUs are one way to convey the facial action, such as six basic facial expressions. AU classifier is used to help generator learn the facial action units' distributions. The loss of AU classifier is defined as
\begin{equation}
L_{cls} = \sum \mathbb{E}_{I_{out} \sim p( I_{out})} \left [ \left\|C_{I}(I_{gt})-C_{I}(I_{out})\right\|_2 \right],
\end{equation}
where $C_I$ is the AU classifier to recognize facial action units. $C_{I}(I_{gt})$ are ground-truth's logits of the last fully connection (FC) layer before activation of AU classifier, $C_{I}(I_{out})$ are generated face's logits of the last FC layer before activation of AU classifier.
The overall loss of the proposed facial expression restoration framework is
\begin{equation}
L_{\text{G}} = L_{pix} + \lambda_1 L_{adv} +\lambda_2 L_{cls} + \lambda_3 L_{per},
\end{equation}
where $\lambda_1$, $\lambda_2$, and $\lambda_3$ are trade-off parameters. General GAN loss for learning discriminator and cross-entropy loss for learning classifier. This max-min game will help to generate realistic face image. Note that for each AU, we should calculate the cross-entropy loss for AU classifier, because it is a multi-label task. The activation function used in this paper is sigmoid function.

\subsection{Improved Graph Convolutional Networks} \label{IGCN}
\begin{figure*}
\centering
\subfloat{\includegraphics[width=0.6 \linewidth]{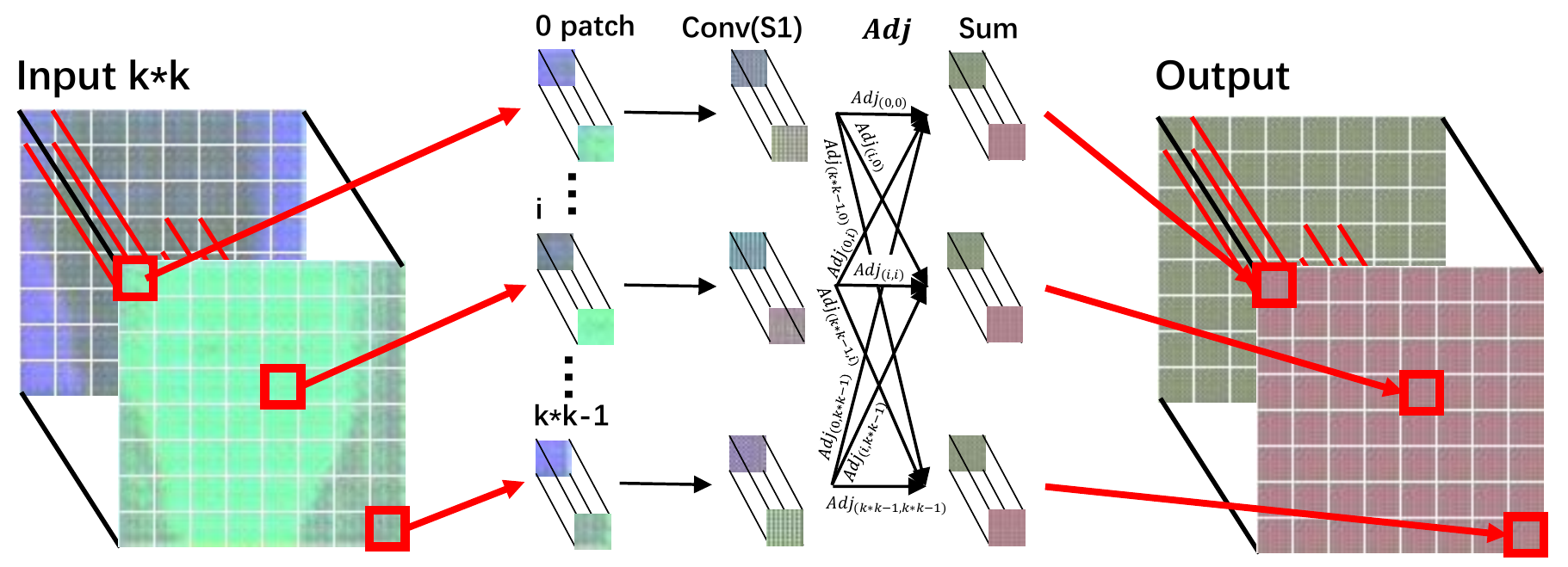}}
\hspace{0mm}
\subfloat{\includegraphics[width=0.35 \linewidth]{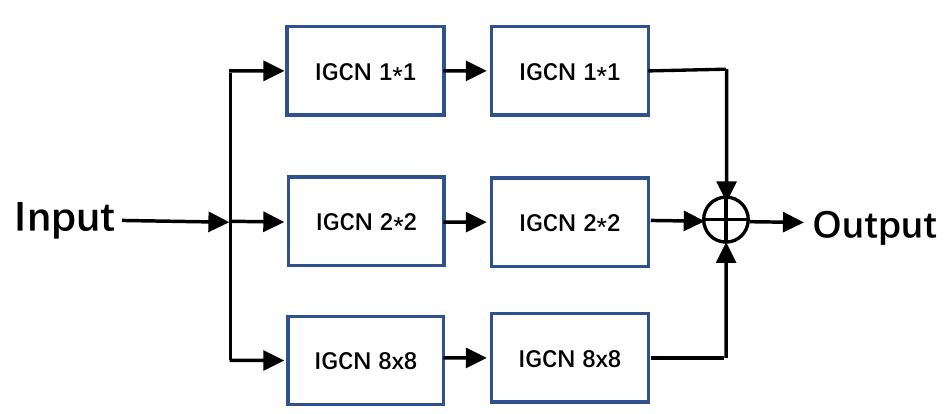}}
\hspace{0mm}
\caption{Structure of the IGCN $8*8$ with stride $1$ in the left. $Adj_{(i,j)}$ represents the link between $i$-th patch and $j$-th patch, $Adj$ represents adjacency matrix. The right is the structure of RRMB, which consists of IGCN $1*1$, $2*2$, $8*8$. $1*1$ represents splitting only 1 patch for the inputs, $2*2$ represents splitting 4 patches for the inputs, and $8*8$ represents splitting 64 patches for the inputs.}
\label{fig:igcn_rrmb}
\vspace{-5mm}
\end{figure*}

The features of the nodes are vectors in conventional graph convolutional network~\cite{kipf2016semi}, which is called as non-euclidean data. For face image, every patch of face image is associated with other patches and is euclidean data. In order to use the face patches as nodes directly via modeling the relations among different facial patches, an improved graph convolutional network is proposed, of which the whole structure is shown in Fig.~\ref{fig:igcn_rrmb}. Due to the ability of graph convolutional network, We can use unoccluded regions to complete the occluded regions via pre-defined relations, such as using unoccluded left eyes to restore occluded right eyes, also can using unoccluded region to enhance the quality of other unoccluded regions. 

Firstly, we split the feature $F$ of face image into $k*k$ face patches with certain order position ID. For each face patch, we use convolutional layer to transform its representation. Here, it should be noted that conventional graph convolutional networks use vectors as features, and use linear transformation layer to capture representations. Different from conventional GCNs, our proposed IGCN uses 4-D tensor of face patches as input features, and uses convolution layer to capture representations. The weights of all convolutional layers for each patch are shared under one layer of IGCN. According to symmetrical adjacency matrix, we get every face patch feature after sum operation. Lastly, we convert $k*k$ face patch features into a feature map according the origin position ID. Note that we also can use deconvolutional layer in IGCN. Adjacency matrix is predefined via facial structure. IGCN can be defined as:
\begin{equation}
    F_{upd} = A:Relu \left( W*F + b \right),
\end{equation}
where F is the stacked patches features, $A$ is the normalized adjacency matrix and $:$ represents tensor product. The adjacency matrixis defined by the correlations of facial structure, such as the symmetry of left face and right face and AUs correlation. Supposing that two patches have relation, then the link between the two patches is suppose to be 1, the opposite is 0. Here we define the relation by cosine similarity between two patches and the co-existence and exclusive relation between two patches consisting of two AUs according landmarks. 

\subsection{Region Relation Modeling Block} \label{RRB}

Region Relation Modeling Block (RRMB) is designed to model the relations of different face patches. This multi-scale structure design is popular in image feature representation learning area. In order to capture different scales' features, we use three scales, such as splitting $1*1$ patch, $2*2$ patches, $8*8$ patches. While splitting $1*1$ patch in IGCN $1*1$, which is same with standard convolutional layer. This scale is to capture global image-level features. The second scale is splitting $2*2$ patches, IGCN $2*2$ is to ensure stable features during flipped situation. This scale setting is exploited to capture object-level features. The third scale is splitting 8*8 patches, IGCN 8*8 is to construct associations between relational spatial patches, such as eyes and mouth. This scale setting is exploited to capture patch-level features. All these scales features are summed pixel-wisely to get the final output features.

\begin{figure*}[t]
    \centering
\subfloat{\includegraphics[width=0.45 \linewidth]{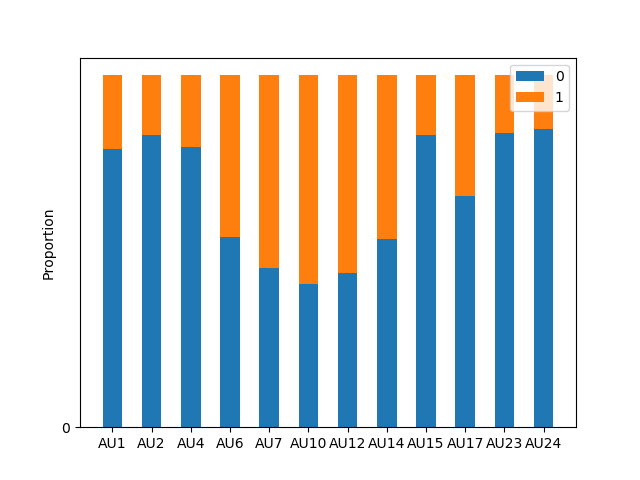}}
\hspace{0mm}
\subfloat{\includegraphics[width=0.45 \linewidth]{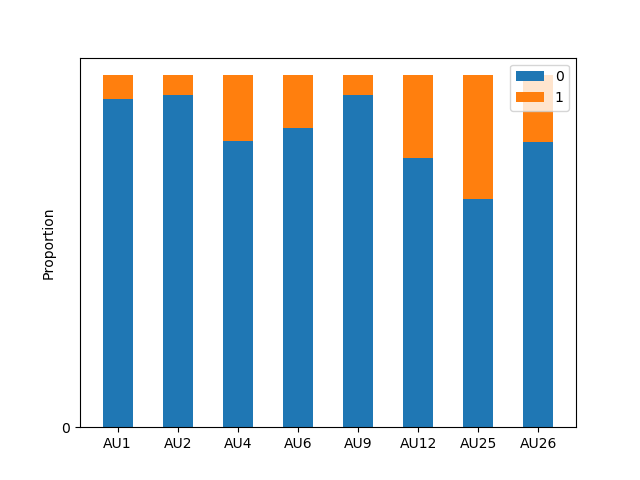}}
\hspace{0mm}
\caption{Sample distributions of BP4D dataset (left) and DISFA dataset (right). 1 represents AU appearance, 0 represents AU disappearance.}
\label{fig:distribution}
\vspace{-5mm}
\end{figure*}
\section{Experiments}
\subsection{Datasets and Settings}
\textbf{Datasets:}
Our proposed facial expression restoration network is evaluated on two wildely used datastes for facial expression analysis: BP4D~\cite{zhang2013high} and DISFA~\cite{mavadati2013disfa}.
The settings of two datasets are similar with \cite{shao2018deep}. For \textbf{BP4D}, we split the dataset to training/testing sets according to subject. There are 28 subjects in training set and 13 subjects in testing set. Each set contains 12 AUs with AU labels of occurrence or absence. Total of 100760 frames are used for training and 45809 frames are used for testing. For \textbf{DISFA}, the processing of the dataset is same as BP4D, there are 18 subjects in training set and 9 subjects in testing set. Each set contains 8 AUs with AU labels of occurrence or absence. Total of 87209 frames are adopted as training set and 43605 frames are used for testing. Note that the color and background of face image are large of differences in DISFA dataset, which is difficult for model to learn well results. The sample distributions of two datasets are shown in Fig.~\ref{fig:distribution}, which illustrates the extreme unbalance situation of labels.

\textbf{Preprocessing:}
 For each face image, we perform similarity transformation including rotation, uniform scaling, and translation to obtain a $128*128*3$ face. This transformation is shape-preserving and brings no change to the expression. The input ill face images are produced by resizing high resolution face image to $16*16$ via bicubic interpolation method and added a random binary mask, of which size is one fourth of the input size.

\textbf{Implementation details:}
We firstly pre-train the AUs classifier, then jointly learn the generator and discriminator, about learning one time of discriminator every three times of generator. For AUs classifier learning, we get the good metrics, which are little lower than the state of the arts. The settings of trade-off parameters are $\lambda_1 = 0.001$, $\lambda_2 = 0.001$, $\lambda_3 = 0.5$. We use Adam for optimization. The learning rate is $0.0001$, the batch size is $8$, and the kernel size is $3$. 

\subsection{Visual results}
\begin{figure*}[t]
    \centering
    \includegraphics[scale=0.3]{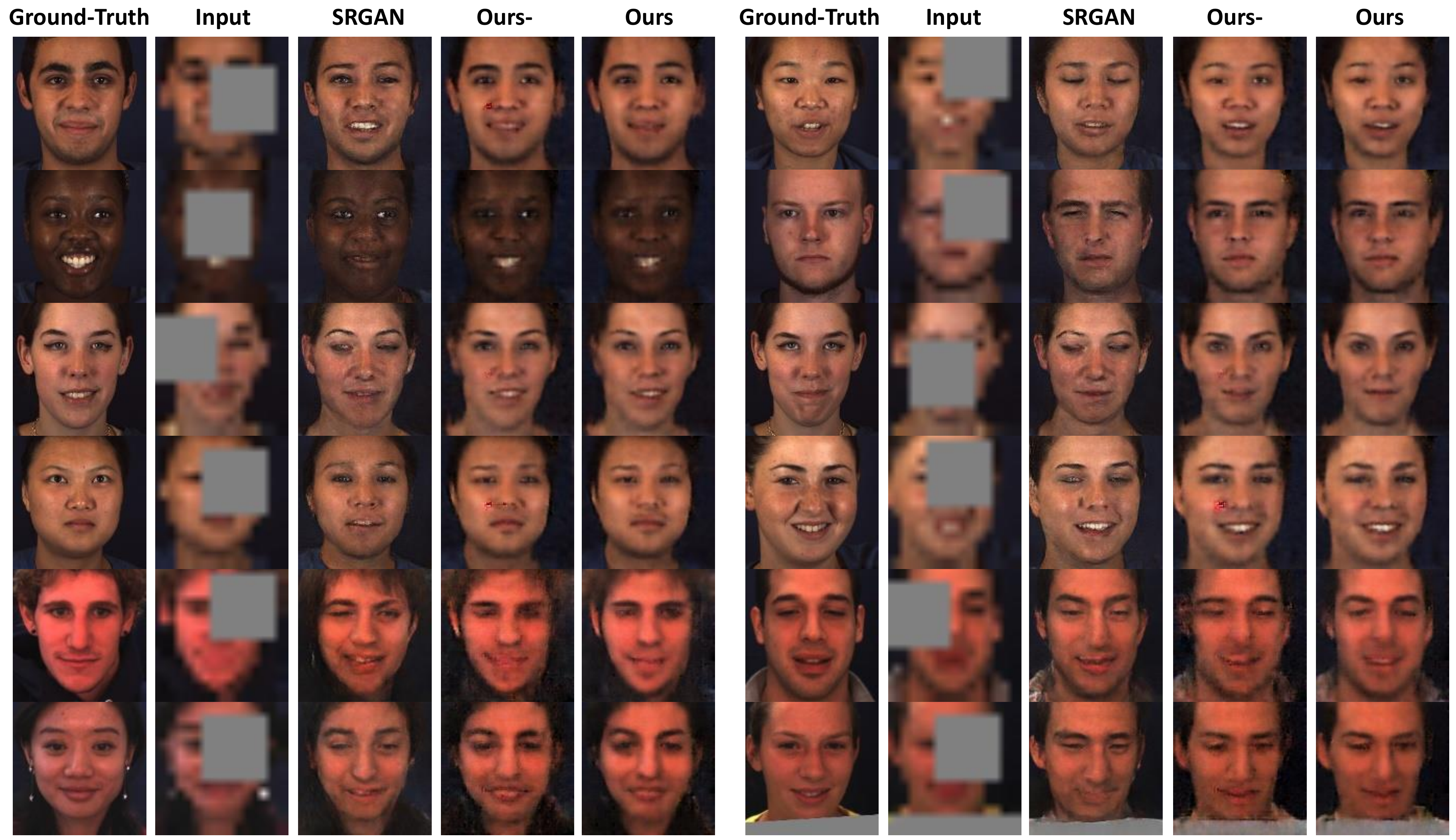}
    \caption{Facial expression restoration results on test datasets, top four rows show the comparison on BP4D and others on DISFA. Zoom in for better view on details.} 
    \label{fig:vis_bp4d}
\vspace{-3mm}
\end{figure*}
We aim to jointly address face inpainting and super-resolution problem for faial expression restoration. General face inpainting methods are not proper to deal with this case, because the numbers of downsample and upsamle layers are same. Here we aim to verify our proposed method's effectiveness in two dataset with SRGAN~\cite{ledig2017photo}, which is notable on image super-resolution area and when we use IGCN $1*1$, the proposed model is similar with residual block\cite{he2016deep}, which is the main component for SRGAN. Comparison with SRGAN and proposed model without AUs detection is shown in Fig.~\ref{fig:vis_bp4d}. In order to observe the difference between different methods, we emphasize the difference of the eye area and mouth area. Obviously, in the results of first row, SRGAN method generates tooth but our proposed method generates closed mouth, which is similar with ground-truth corresponding to AU 25, and in the results of third row, SRGAN generates closed eyes but our proposed method generates opened eyes corresponding to AU 43. For quality and reality, the results of SRGAN have virtual streak and blur, such as the first row. It is worth noting that the shown results are from 13 subjects of  BP4D dataset and 9 subjects of DISFA dataset for testing respectively. It is observed that our proposed method outperforms SRGAN in the aspects of reality and quality. And also, our proposed method can well retain the facial action or expression after restoration.

\begin{table*}[!t]
\centering
\caption{F1-score and accuracy for 12 AUs on BP4D. Ours is the total loss to learn the proposed model, ours- lacks the loss of AUs classifier.}
\vspace{1mm}
\scalebox{0.73}{
\begin{tabular}{c|c|c|c|c|c|c|c|c}
\hline
    & \multicolumn{4}{c|}{F1-score} & \multicolumn{4}{c}{Accuracy}   \\ \cline{2-9}
\textbf{BP4D} & SRGAN & Ours- & Ours & Ground-Truth & SRGAN & Ours- & Ours & Ground-Truth \\ \hline
1  &$0.097$  &$0.349$  &$0.412$    &$0.412$  &$0.700$  &$0.690$  &$0.674$  & $0.720$ \\
2  &$0.065$  &$0.214$  &$0.267$    &$0.331$  &$0.754$  &$0.770$  &$0.749$  & $0.725$ \\
4  &$0.168$  &$0.284$  &$0.298$    &$0.276$  &$0.692$  &$0.798$  &$0.803$  & $0.785$ \\
6  &$0.545$  &$0.575$  &$0.672$    &$0.730$  &$0.503$  &$0.705$  &$0.731$  & $0.732$ \\
7  &$0.560$  &$0.619$  &$0.594$    &$0.646$  &$0.500$  &$0.662$  &$0.650$  & $0.646$ \\
10 &$0.680$  &$0.768$  &$0.905$    &$0.825$  &$0.565$  &$0.748$  &$0.775$  & $0.777$ \\
12 &$0.616$  &$0.882$  &$0.861$    &$0.874$  &$0.528$  &$0.840$  &$0.837$  & $0.845$ \\
14 &$0.569$  &$0.676$  &$0.594$    &$0.634$  &$0.494$  &$0.606$  &$0.587$  & $0.595$ \\
15 &$0.177$  &$0.258$  &$0.299$    &$0.307$  &$0.712$  &$0.713$  &$0.792$  & $0.784$ \\
17 &$0.446$  &$0.748$  &$0.723$    &$0.548$  &$0.464$  &$0.597$  &$0.626$  & $0.595$ \\
23 &$0.204$  &$0.433$  &$0.454$    &$0.333$  &$0.652$  &$0.755$  &$0.744$  & $0.684$ \\
24 &$0.142$  &$0.080$  &$0.086$    &$0.411$  &$0.753$  &$0.763$  &$0.832$  & $0.794$ \\ \hline
Avg &$0.356$  &$0.491$  &$0.514$    &$0.525$  &$0.610$  &$0.721$  &$0.733$  & $0.724$ \\ \hline



\end{tabular}}
\label{tb:bp4d}
\vspace{-2mm}
\end{table*}
\begin{table*}[!t]
\centering
\caption{F1-score and accuracy for 8 AUs on DISFA. Ours is the total loss to learn the proposed model, ours- lacks the loss of AUs classifier.}
\vspace{1mm}
\scalebox{0.73}{
\begin{tabular}{c|c|c|c|c|c|c|c|c}
\hline
    & \multicolumn{4}{c|}{F1-score} & \multicolumn{4}{c}{Accuracy}   \\ \cline{2-9}
\textbf{DISFA} & SRGAN & Ours- & Ours & Ground-Truth & SRGAN & Ours- & Ours & Ground-Truth \\ \hline
1  &$0.107$  &$0.217$  &$0.214$    &$0.219$  &$0.901$  &$0.863$  &$0.866$  & $0.908$ \\
2  &$0.010$  &$0.159$  &$0.185$    &$0.198$  &$0.958$  &$0.806$  &$0.836$  & $0.949$ \\
4  &$0.091$  &$0.202$  &$0.217$    &$0.286$  &$0.739$  &$0.739$  &$0.716$  & $0.735$ \\
6  &$0.392$  &$0.336$  &$0.348$    &$0.362$  &$0.870$  &$0.797$  &$0.804$  & $0.855$ \\
9  &$0.039$  &$0.185$  &$0.365$    &$0.270$  &$0.903$  &$0.945$  &$0.939$  & $0.885$ \\
12 &$0.576$  &$0.548$  &$0.558$    &$0.628$  &$0.763$  &$0.741$  &$0.752$  & $0.844$ \\
25 &$0.509$  &$0.654$  &$0.765$    &$0.746$  &$0.608$  &$0.809$  &$0.811$  & $0.772$ \\
26 &$0.270$  &$0.506$  &$0.504$    &$0.524$  &$0.685$  &$0.725$  &$0.723$  & $0.755$ \\ \hline
Avg &$0.249$  &$0.351$  &$0.395$    &$0.404$  &$0.803$  &$0.803$  &$0.806$  & $0.838$ \\ \hline

\end{tabular}}
\label{tb:disfa}
\vspace{-5mm}
\end{table*}

\subsection{Quantity results}
In order to investigate the effectiveness of AU classifier in our framework for facial expression restoration, Table.~\ref{tb:bp4d} and Table.~\ref{tb:disfa} present the AU detection results on our restored facial expression images compared with corresponding ground truth images in terms of F1-score and accuracy, where "$Ours-$" is the proposed framework without AU classifier. Note that AU classifier is trained on training dataset but the quantity results are compared in results of different model on test dataset.  

The results shown in Table.~\ref{tb:bp4d} demonstrates that our proposed method outperforms SRGAN in BP4D dataset, even "$Ours-$" brings significant increments of $13.6\%$ and $11.1\%$ respectively for average f1-score and accuracy than SRGAN. It is also observed that learning framework with AU classifier gain a few increments of $2.3\%$ and $1.2\%$ for average f1-score and accuracy. The large gaps between our proposed method and SRGAN are associated with the distribution of BP4D. Due to the unbalanced distribution of status for each AU as is shown in Fig.~\ref{fig:distribution}, our proposed facial expression restoration method is inclined to learn the `0' status for each AU. For accuracy, ours result is similar with ground-truth, even better than ground-truth. And the AU classifier is strong to make the logits of generated face image incline to the distribution of the AUs, which results in the our proposed model getting little higher accuracy than ground-truth. 

Similar results in DISFA dataset are shown in Table.~\ref{tb:disfa}, from which it can be observed that our proposed method with or without AUs classifier outperforms SRGAN. Specifically, our proposed method increase $14.6\%$ and $0.3\%$ for average f1-score and accuracy than SRGAN. For accuracy, our proposed method has a few improvements, it is also associated with AUs occur distribution in the dataset. For most face images, AUs do not occur in DISFA dataset as shown in Fig.~\ref{fig:distribution}, so AU classifier always recognizes `0' status. On the other hand, lower f1-score also tell us that model learns the nature situation easier than activate situation for each AU. 

\begin{table}[t!]
\centering
\caption{SSIM and PSNR on BP4D and DISFA.}
\vspace{1mm}
\scalebox{0.73}{
\begin{tabular}{c|c|c|c|c|c|c}
\hline
    & \multicolumn{3}{c|}{SSIM} & \multicolumn{3}{c}{PSNR}   \\ \cline{2-7}
 & SRGAN & Ours- & Ours  & SRGAN & Ours- & Ours  \\ \hline
BP4D  &$0.625$  &$0.738$  &$0.748$    &$23.324$  &$25.297$  &$25.418$   \\ \hline
DISFA  &$0.598$  &$0.624$  &$0.655$    &$20.855$  &$21.621$  &$21.826$  \\ \hline
\end{tabular}}
\label{tb:ssim}
\vspace{-6mm}
\end{table}

When it comes to image restoration task, many works often compare the structural similarity(SSIM) and Peak Signal to Noise Ratio(PSNR). The results can be observed in Table.~\ref{tb:ssim}. Our proposed method outperforms $0.123$ and $0.057$ for SSIM, $2.094$ and $0.971$ for PSNR respectively in BP4D and DISFA dataset, which demonstrate the effectiveness of our proposed method on facial expression restoration. Note that there are few improvements in DISFA datasets due to its extreme unbalance distribution.

\section{Conclusion}
In this paper, we have proposed a novel facial expression restoration method by integrating a region relation modeling block with the aid of an improved graph convolution network to model the relations among different facial regions. The proposed method is beneficial to facial expression analysis under challenging environments, i.e. low resolution and occlusion. Extensive qualitative and quantitative evaluations conducted on BP4D and DISFA have demonstrated the effectiveness of our method for facial expression restoration. The proposed framework is also promising to be applied for other face restoration tasks and other multi-task problems, i.e. face recognition, facial attribute analysis, etc.

\vspace{-3mm}
\section*{Acknowledgements}
\vspace{-3mm}
This work is supported by the National Natural Science Foundation of China under Grants of 41806116 and 61503277. We gratefully acknowledge the support of NVIDIA Corporation with the donation of the Titan V GPU used for this research.

%

\bibliography{egbib}

\begin{thebibliography}{10}
\providecommand{\url}[1]{\texttt{#1}}
\providecommand{\urlprefix}{URL }
\providecommand{\doi}[1]{https://doi.org/#1}

\bibitem{chen2018fsrnet}
Chen, Y., Tai, Y., Liu, X., Shen, C., Yang, J.: Fsrnet: End-to-end learning
  face super-resolution with facial priors. In: Proceedings of the IEEE
  Conference on Computer Vision and Pattern Recognition. pp. 2492--2501 (2018)

\bibitem{cheon2018generative}
Cheon, M., Kim, J.H., Choi, J.H., Lee, J.S.: Generative adversarial
  network-based image super-resolution using perceptual content losses. In:
  Proceedings of the European Conference on Computer Vision (ECCV). pp.~0--0
  (2018)

\bibitem{ekman1997face}
Ekman, R.: What the face reveals: Basic and applied studies of spontaneous
  expression using the Facial Action Coding System (FACS). Oxford University
  Press, USA (1997)

\bibitem{engin2018cycle}
Engin, D., Gen{\c{c}}, A., Kemal~Ekenel, H.: Cycle-dehaze: Enhanced cyclegan
  for single image dehazing. In: Proceedings of the IEEE Conference on Computer
  Vision and Pattern Recognition Workshops. pp. 825--833 (2018)

\bibitem{gilmer2017neural}
Gilmer, J., Schoenholz, S.S., Riley, P.F., Vinyals, O., Dahl, G.E.: Neural
  message passing for quantum chemistry. In: Proceedings of the 34th
  International Conference on Machine Learning-Volume 70. pp. 1263--1272. JMLR.
  org (2017)

\bibitem{goodfellow2014generative}
Goodfellow, I., Pouget-Abadie, J., Mirza, M., Xu, B., Warde-Farley, D., Ozair,
  S., Courville, A., Bengio, Y.: Generative adversarial nets. In: Advances in
  neural information processing systems. pp. 2672--2680 (2014)

\bibitem{he2016deep}
He, K., Zhang, X., Ren, S., Sun, J.: Deep residual learning for image
  recognition. In: Proceedings of the IEEE conference on computer vision and
  pattern recognition. pp. 770--778 (2016)

\bibitem{huang2017beyond}
Huang, R., Zhang, S., Li, T., He, R.: Beyond face rotation: Global and local
  perception gan for photorealistic and identity preserving frontal view
  synthesis. In: Proceedings of the IEEE International Conference on Computer
  Vision. pp. 2439--2448 (2017)

\bibitem{khorrami2015deep}
Khorrami, P., Paine, T., Huang, T.: Do deep neural networks learn facial action
  units when doing expression recognition? In: Proceedings of the IEEE
  International Conference on Computer Vision Workshops. pp. 19--27 (2015)

\bibitem{kipf2016semi}
Kipf, T.N., Welling, M.: Semi-supervised classification with graph
  convolutional networks. arXiv preprint arXiv:1609.02907  (2016)

\bibitem{krizhevsky2012imagenet}
Krizhevsky, A., Sutskever, I., Hinton, G.E.: Imagenet classification with deep
  convolutional neural networks. In: Advances in neural information processing
  systems. pp. 1097--1105 (2012)

\bibitem{ledig2017photo}
Ledig, C., Theis, L., Husz{\'a}r, F., Caballero, J., Cunningham, A., Acosta,
  A., Aitken, A., Tejani, A., Totz, J., Wang, Z., et~al.: Photo-realistic
  single image super-resolution using a generative adversarial network. In:
  Proceedings of the IEEE conference on computer vision and pattern
  recognition. pp. 4681--4690 (2017)

\bibitem{li2018learning}
Li, L., Pan, J., Lai, W.S., Gao, C., Sang, N., Yang, M.H.: Learning a
  discriminative prior for blind image deblurring. In: Proceedings of the IEEE
  Conference on Computer Vision and Pattern Recognition. pp. 6616--6625 (2018)

\bibitem{li2017eac}
Li, W., Abtahi, F., Zhu, Z., Yin, L.: Eac-net: A region-based deep enhancing
  and cropping approach for facial action unit detection. In: 2017 12th IEEE
  International Conference on Automatic Face \& Gesture Recognition (FG 2017).
  pp. 103--110. IEEE (2017)

\bibitem{li2017generative}
Li, Y., Liu, S., Yang, J., Yang, M.H.: Generative face completion. In:
  Proceedings of the IEEE Conference on Computer Vision and Pattern
  Recognition. pp. 3911--3919 (2017)

\bibitem{mavadati2013disfa}
Mavadati, S.M., Mahoor, M.H., Bartlett, K., Trinh, P., Cohn, J.F.: Disfa: A
  spontaneous facial action intensity database. IEEE Transactions on Affective
  Computing  \textbf{4}(2),  151--160 (2013)

\bibitem{mirza2014conditional}
Mirza, M., Osindero, S.: Conditional generative adversarial nets. arXiv
  preprint arXiv:1411.1784  (2014)

\bibitem{muhammad2018image}
Muhammad, N., Bibi, N., Jahangir, A., Mahmood, Z.: Image denoising with norm
  weighted fusion estimators. Pattern Analysis and Applications
  \textbf{21}(4),  1013--1022 (2018)

\bibitem{peng2018weakly}
Peng, G., Wang, S.: Weakly supervised facial action unit recognition through
  adversarial training. In: Proceedings of the IEEE Conference on Computer
  Vision and Pattern Recognition. pp. 2188--2196 (2018)

\bibitem{shao2018deep}
Shao, Z., Liu, Z., Cai, J., Ma, L.: Deep adaptive attention for joint facial
  action unit detection and face alignment. In: Proceedings of the European
  Conference on Computer Vision (ECCV). pp. 705--720 (2018)

\bibitem{simonyan2014very}
Simonyan, K., Zisserman, A.: Very deep convolutional networks for large-scale
  image recognition. arXiv preprint arXiv:1409.1556  (2014)

\bibitem{song2018joint}
Song, Y., Zhang, J., Gong, L., He, S., Bao, L., Pan, J., Yang, Q., Yang, M.H.:
  Joint face hallucination and deblurring via structure generation and detail
  enhancement. International Journal of Computer Vision pp. 1--16 (2018)

\bibitem{taheri2014structure}
Taheri, S., Qiu, Q., Chellappa, R.: Structure-preserving sparse decomposition
  for facial expression analysis. IEEE Transactions on Image Processing
  \textbf{23}(8),  3590--3603 (2014)

\bibitem{wang2018stacked}
Wang, J., Li, X., Yang, J.: Stacked conditional generative adversarial networks
  for jointly learning shadow detection and shadow removal. In: Proceedings of
  the IEEE Conference on Computer Vision and Pattern Recognition. pp.
  1788--1797 (2018)

\bibitem{wang2018rain}
Wang, Y.T., Zhao, X.L., Jiang, T.X., Deng, L.J., Chang, Y., Huang, T.Z.: Rain
  streak removal for single image via kernel guided cnn. arXiv preprint
  arXiv:1808.08545  (2018)

\bibitem{wu2016constrained}
Wu, Y., Ji, Q.: Constrained joint cascade regression framework for simultaneous
  facial action unit recognition and facial landmark detection. In: Proceedings
  of the IEEE conference on computer vision and pattern recognition. pp.
  3400--3408 (2016)

\bibitem{yeh2017semantic}
Yeh, R.A., Chen, C., Yian~Lim, T., Schwing, A.G., Hasegawa-Johnson, M., Do,
  M.N.: Semantic image inpainting with deep generative models. In: Proceedings
  of the IEEE Conference on Computer Vision and Pattern Recognition. pp.
  5485--5493 (2017)

\bibitem{zhang2013high}
Zhang, X., Yin, L., Cohn, J.F., Canavan, S., Reale, M., Horowitz, A., Liu, P.:
  A high-resolution spontaneous 3d dynamic facial expression database. In: 2013
  10th IEEE International Conference and Workshops on Automatic Face and
  Gesture Recognition (FG). pp.~1--6. IEEE (2013)

\bibitem{zhao2015joint}
Zhao, K., Chu, W.S., De~la Torre, F., Cohn, J.F., Zhang, H.: Joint patch and
  multi-label learning for facial action unit detection. In: Proceedings of the
  IEEE Conference on Computer Vision and Pattern Recognition. pp. 2207--2216
  (2015)

\bibitem{zhao2016deep}
Zhao, K., Chu, W.S., Zhang, H.: Deep region and multi-label learning for facial
  action unit detection. In: Proceedings of the IEEE Conference on Computer
  Vision and Pattern Recognition. pp. 3391--3399 (2016)

\end{thebibliography}
\bibliographystyle{splncs04}

\end{document}